\documentclass[]{article}
\usepackage[letterpaper]{geometry}
\usepackage{mtsummit2017}
\usepackage{url}
\usepackage{times}
\usepackage{natbib}
\usepackage{layout}

\usepackage{latexsym}
\usepackage{epsfig}
\usepackage{multirow}
\usepackage{graphicx}
\usepackage{amssymb}
\usepackage{amsmath}
\usepackage{amsfonts}
\usepackage{subfigure}
\usepackage{color}

\begin{document}

\title{\bf Neural Machine Translation Model\\ 
with a Large Vocabulary Selected \\
by Branching Entropy}

\author{\name{\bf Zi Long} \hfill \\
        \name{\bf Ryuichiro Kimura} \hfill \\
        \name{\bf Takehito Utsuro} \hfill \\
        \addr{Grad. Sc. Sys. \& Inf. Eng., University of Tsukuba,
        tsukuba, 305-8573, Japan}
\AND
        \name{\bf Tomoharu Mitsuhashi} \hfill \\
        \addr{Japan Patent Information Organization, 4-1-7, Tokyo,
        Koto-ku, Tokyo, 135-0016, Japan}
\AND
        \name{\bf Mikio Yamamoto} \hfill \\
        \addr{Grad. Sc. Sys. \& Inf. Eng., University of Tsukuba,
        tsukuba, 305-8573, Japan}
}

\maketitle
\pagestyle{empty}

\begin{abstract}
 Neural machine translation (NMT), a new approach to machine translation,
 has achieved promising results comparable to those of traditional
 approaches such as statistical machine translation (SMT). Despite its
 recent success, NMT cannot handle a larger vocabulary because the
 training complexity and decoding complexity proportionally increase
 with the number of target words. This problem becomes even more serious
 when translating patent documents, which contain many technical terms
 that are observed infrequently. In this paper, we propose to select
 phrases that contain out-of-vocabulary words using the
 statistical approach of branching entropy. This allows the proposed NMT
 system to be applied to a translation task of any language pair without
 any language-specific knowledge about technical term
 identification. The selected phrases are then replaced with
 tokens during training and post-translated by the phrase translation
 table of SMT. Evaluation on Japanese-to-Chinese, 
 Chinese-to-Japanese, Japanese-to-English and English-to-Japanese
 patent sentence translation proved the effectiveness of phrases 
 selected with branching entropy, where the proposed NMT model achieves a
 substantial improvement over a baseline NMT model without our
 proposed technique. 
 Moreover, the number of translation errors of under-translation by the
 baseline NMT model without our proposed technique reduces to around
 half by the proposed NMT model. 

\end{abstract}

\section{Introduction}
\label{sec:inro}

\begin{figure*}
   \centering
   \includegraphics[scale=0.43]{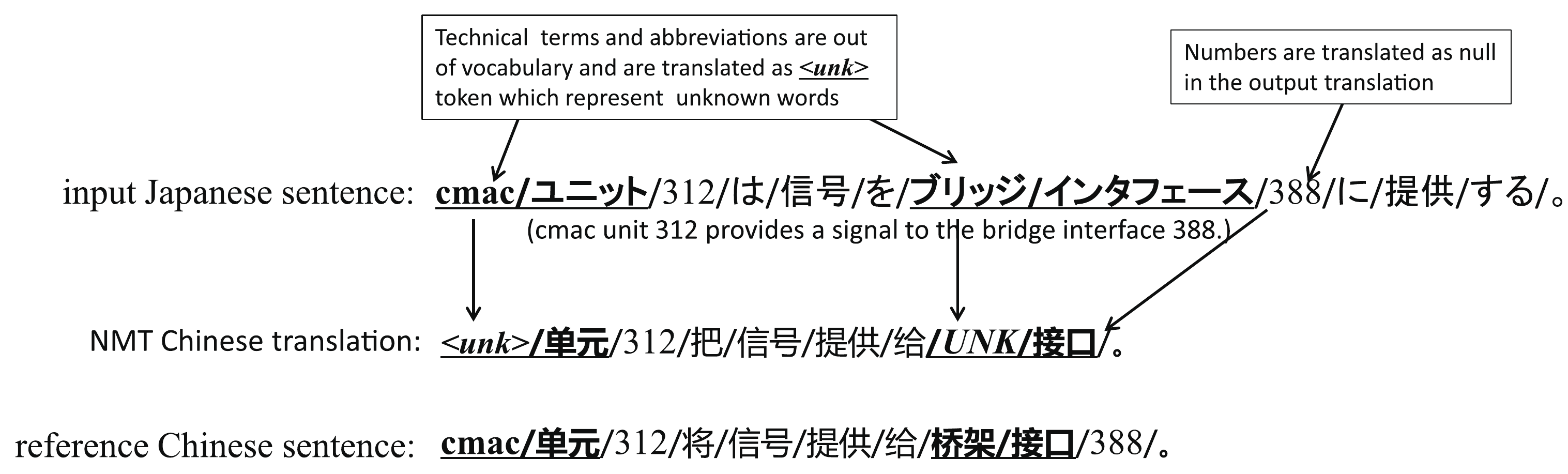}
   \caption{Example of translation errors when translating patent
   sentences with technical terms using NMT}
   \label{fig:problem}
  \end{figure*}

  Neural machine translation (NMT), a new approach to solving machine
  translation, has achieved promising results
  \citep{Bahdanau15a,Cho14a,Jean15a,Kalch13a,Luong15b,Luong15a,Sutskever14a}.
  An NMT system builds a simple large neural network that reads the
  entire input source sentence and generates an output translation. The
  entire neural network is jointly trained to maximize the conditional
  probability of the correct translation of a source sentence with a
  bilingual corpus. Although NMT offers many advantages over traditional
  phrase-based approaches, such as a small memory footprint and simple
  decoder implementation, conventional NMT is limited when it comes to
  larger vocabularies. This is because the training complexity and
  decoding complexity proportionally increase with the number of target
  words. Words that are out of vocabulary are represented by a single
  ``$\langle unk \rangle$'' token in translations, as illustrated in Figure~\ref{fig:problem}.
  The problem becomes more serious when translating patent documents, which contain
  several newly introduced technical terms.
  
  There have been a number of related studies that address the
  vocabulary limitation of NMT systems.
  \cite{Jean15a} provided  
  an efficient approximation to the softmax function to accommodate a very large
  vocabulary in an NMT system. 
  \cite{Luong15a} proposed annotating the occurrences of
  the out-of-vocabulary token in the target sentence 
  with positional information to track its
  alignments, after which they replace the tokens
  with their translations using simple word dictionary lookup or
  identity copy. \cite{XiaLi16a} proposed replacing out-of-vocabulary 
  words with similar in-vocabulary words 
  based on a similarity model learnt from monolingual data.
  \cite{Sennrich16a} introduced an effective approach based on encoding  
  rare and out-of-vocabulary words as sequences of subword units.
  \cite{Luong16a} provided a character-level and word-level hybrid NMT
  model to achieve an open vocabulary, and \cite{Costa-Jussa16a} proposed 
  an NMT system that uses character-based embeddings.

  However, these previous approaches have limitations when translating
  patent sentences. This is because their methods only focus on
  addressing the problem of out-of-vocabulary 
  words even though the words are
  parts of technical terms. It is obvious that a technical term should be
  considered as one word that comprises 
  components that always have different meanings and translations when
  they are used alone. An example is shown in Figure~\ref{fig:problem},
  where the Japanese word 
  ``\includegraphics[scale=0.45]{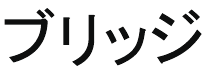}''(bridge) 
  should be translated to Chinese word
  ``\includegraphics[scale=0.45]{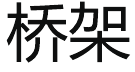}'' 
  when included in technical term ``bridge interface''; however, it is
  always translated as ``\includegraphics[scale=0.45]{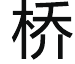}''.

  To address this problem, \cite{Long16b} proposed
  extracting compound nouns as technical terms and replacing them with
  tokens. These compound nouns 
  then are post-translated with the phrase translation table of the
  statistical machine translation (SMT) system. 
  However, in their work
  on Japanese-to-Chinese patent translation, Japanese compound nouns are
  identified using several heuristic rules that use
  specific linguistic knowledge based on part-of-speech tags of
  morphological analysis of Japanese language,
  and thus, the NMT system has limited
  application to the translation task of other language pairs. In this
  paper, based on the approach of training an NMT model on a bilingual
  corpus wherein technical term pairs are replaced with tokens as in
  \cite{Long16b}, we aim to select phrase pairs using the
  statistical approach of branching entropy; this allows the proposed
  technique to be applied to the translation task on any language pair
  without needing specific language knowledge to formulate the rules for
  technical term identification. 
  Based on the results of our experiments
  on many pairs of languages: Japanese-to-Chinese, Chinese-to-Japanese,
  Japanese-to-English and English-to-Japanese, the proposed NMT model
  achieves a substantial improvement over a baseline NMT model without our
  proposed technique. Our proposed NMT model achieves 
  an improvement of 1.2 BLEU points over a baseline NMT model when
  translating Japanese sentences into Chinese, and an improvement of 1.7
  BLEU points when translating Chinese sentences into Japanese. Our
  proposed NMT model achieves an improvement of 1.1 BLEU points over a
  baseline NMT model when translating Japanese sentences into English,
  and an improvement of 1.4 BLEU points when translating English
  sentences into Japanese. 
  Moreover, the number of translation error of 
  under-translations\footnote{
    It is known that NMT models tend to have
    the problem of the under-translation.
    \cite{Tu16a} proposed coverage-based NMT
    which considers the problem of the under-translation.
  } 
  by the the baseline NMT model without our proposed technique reduces
  to around half by the proposed NMT model.

\section{Neural Machine Translation}
\label{sec:nmt}

 NMT uses a single
 neural network trained jointly to maximize the translation performance
 ~\citep{Bahdanau15a,Cho14a,Kalch13a,Luong15b,Sutskever14a}.
 Given a source sentence {\boldmath $x$} $=(x_1,\ldots,x_N)$ and target
 sentence {\boldmath $y$} $=(y_1,\ldots,y_M)$, an NMT model uses a neural network to
 parameterize the conditional distributions
 \begin{eqnarray}
  p(y_z \mid y_{< z},\mbox{\boldmath $x$}) \nonumber
 \end{eqnarray}
 for $1 \leq z \leq M$. Consequently, it becomes possible to compute and
 maximize the log probability of the target sentence given the source
 sentence as 
 \begin{eqnarray}
  \label{eq:prob}
  \log p(\mbox{\boldmath $y$} \mid \mbox{\boldmath $x$}) =
  \sum_{l=1}^{M} \log p(y_z|y_{< z},\mbox{\boldmath $x$}) \nonumber
 \end{eqnarray}

 In this paper, we use an NMT model similar to that used by \cite{Bahdanau15a},
  which consists of an encoder of a bidirectional long short-term memory
  (LSTM)~\citep{Hochreiter97a} and another LSTM as decoder.
  In the model of \cite{Bahdanau15a}, 
  the encoder consists of forward and backward LSTMs. The forward LSTM
  reads the source sentence as it is 
  ordered (from $x_1$ to $x_N$) and calculates a sequence of forward
  hidden states, while the backward LSTM reads the source sentence in the
  reverse order (from $x_N$ to $x_1$) , resulting in a sequence of
  backward hidden states. 
  The decoder then predicts target words using not only a recurrent hidden
  state and the previously predicted word but also a context vector as followings:
  \begin{eqnarray}
   p(y_z \mid y_{< z},\mbox{\boldmath $x$}) = g(y_{z-1}, s_{z-1}, c_z) \nonumber
  \end{eqnarray}
  where $s_{z-1}$ is an LSTM hidden state of decoder, and $c_z$ is a
  context vector computed from both of the forward hidden states and
  backward hidden states, for $1 \leq z \leq M$.
  
\section{Phrase Pair Selection using Branching Entropy}
\label{sec:extract}
 {\color{black}
 Branching entropy has been applied to
 the procedure of text segmentation (e.g., \citep{Jin06a}) and key phrases
 extraction (e.g., \citep{YChen10a}).
 }
 In this work, we use the left/right branching entropy to detect the
 boundaries  
 of phrases, and thus select phrase pairs automatically. 

\subsection{{\color{black}Branching Entropy}}
\label{sec:bra_entr}

 The left branching entropy and right branching entropy of a phrase
 \mbox{\boldmath $w$} are respectively defined as 
 \begin{eqnarray}
  H_{l}(\mbox{\boldmath $w$}) = - \sum_{v \in V_l^{\mbox{\boldmath $w$}}} p_l(v) \log_2 p_l(v) \nonumber \\
  H_{r}(\mbox{\boldmath $w$}) = - \sum_{v \in V_r^{\mbox{\boldmath $w$}}} p_r(v) \log_2 p_r(v) \nonumber
 \end{eqnarray}
 where \mbox{\boldmath $w$} is the phrase of interest (e.g.,
 ``\includegraphics[scale=0.45]{figures/brigde_ja.eps}/\includegraphics[scale=0.45]{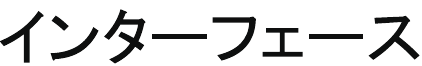}''
 in the Japanese sentence shown in Figure~\ref{fig:problem}, which means
 ``bridge interface''), $V_l^{\mbox{\boldmath $w$}}$ is a set of words that are 
 adjacent to the left of \mbox{\boldmath $w$} (e.g.,
 ``\includegraphics[scale=0.45]{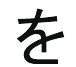}'' in
 Figure~\ref{fig:problem}, which is a Japanese particle) 
 and $V_r^{\mbox{\boldmath $w$}}$ is a set of words that are
 adjacent to the right of \mbox{\boldmath $w$} (e.g., ``388'' in
 Figure~\ref{fig:problem}).  
 The probabilities $p_l(v)$ and $p_r(v)$ are respectively computed as 
\[
  p_l(v)\ =\ \frac{f_{v,\mbox{\boldmath $w$}}}{f_{\mbox{\boldmath $w$}}}
  \ \ \ \ \ \ \ \ 
  p_r(v)\ =\ \frac{f_{\mbox{\boldmath $w$},v}}{f_{\mbox{\boldmath $w$}}} 
\]
 where $f_{\mbox{\boldmath $w$}}$ is the frequency count of phrase
 \mbox{\boldmath $w$}, and $f_{v,\mbox{\boldmath $w$}}$ and
 $f_{\mbox{\boldmath $w$},v}$ are the frequency counts of sequence
 ``$v$,\mbox{\boldmath $w$}'' and sequence ``\mbox{\boldmath $w$},$v$''
 respectively. According to the definition of branching entropy, when a
 phrase $\mbox{\boldmath $w$}$ is a technical term
 that is always used as a compound word, both its left branching
 entropy $H_{l}($\mbox{\boldmath $w$}$)$ and right branching entropy
 $H_{r}($\mbox{\boldmath $w$}$)$ have high values because many different
 words, such as particles and numbers, can be adjacent to the
 phrase. However, the left/right branching entropy of substrings of
 \mbox{\boldmath $w$} have low values because words contained in
 \mbox{\boldmath $w$} are always adjacent to each other.

\subsection{{\color{black}Selecting Phrase Pairs}}
\label{sec:select_entr}

  Given a parallel sentence pair $\langle S_s, S_t\rangle$,
  all $n$-grams phrases of source sentence $S_s$ and target sentence $S_t$
  are extracted and aligned using phrase translation table and word alignment of SMT
  according to the approaches described in \cite{Long16b}.
  Next, phrase translation pair $\langle t_s,
  t_t \rangle$ obtained from $\langle S_s, S_t\rangle$ that satisfies
  all the following conditions is selected as 
  a phrase pair and is extracted:
  \begin{enumerate}
   \item[(1)] \label{item:oov} Either $t_s$ or $t_t$ contains at least one out-of-vocabulary
	      word.\footnote{
	      {\color{black}
	      One of the major focus of this paper is the comparison between the proposed
	      method and \cite{Luong15a}. Since \cite{Luong15a} proposed to pre-process and
	      post-translate only out-of-vocabulary words, we focus only on compound
	      terms which include at least one 
	      out-of-vocabulary words. 
	      }
	      }
   \item[(2)] \label{item:stop}Neither $t_s$ nor $t_t$ contains predetermined stop
	      words.
   \item[(3)] \label{item:entropy}Entropies $H_l(t_s)$, $H_l(t_t)$, $H_r(t_s)$ and $H_r(t_t)$ are
	      larger than a lower bound, while the left/right branching
	      entropy of the substrings of $t_s$ and $t_t$ are lower than or equal
	      to the lower bound.
  \end{enumerate}
  Here, the maximum length of a phrase as well as the lower bound of the branching
  entropy are tuned with the validation set.\footnote{
   Throughout the evaluations on patent translation of both
   language pairs of Japanese-Chinese and Japanese-English, 
   the maximum length of the extracted phrases is tuned as 7.
   The lower bounds of the branching entropy are tuned as
   5 for patent translation of the language pair of Japanese-Chinese,
   and 8 for patent translation of the language pair of
   Japanese-English. 
   We also tune the number of stop words using the validation set, 
   and use the 200 most-frequent Japanese morphemes and Chinese words
   as stop words for the language pair of Japanese-Chinese, 
   use the 100 most-frequent Japanese morphemes and English 
   words as stop words for the language pair of Japanese-English.
  } 
  All the selected source-target phrase pairs are then used in the
  next section as phrase pairs.\footnote{
    We sampled 200 Japanese-Chinese sentence pairs, manually annotated
    compounds and evaluated the approach of phrase extraction with
    the branching entropy. Based on the result,
    (a) 25\% of them are correct, (b) 20\% subsume correct compounds as
    their substrings, (c) 18\% are substrings of correct compounds, (d)
    22\% subsume substrings of correct compounds but other than (b) nor
    (c), and (e) the remaining 15\% are error strings such as functional
    compounds and fragmental strings consisting of numerical expressions.
  }

\section{NMT with a Large Phrase Vocabulary}
\label{sec:trans}

  In this work, the NMT model is trained on a bilingual
  corpus in which phrase pairs are replaced with tokens. The NMT
  system is then used as a decoder to translate the source sentences and
  replace the tokens with phrases translated using SMT.

  \begin{figure*}
   \center
   \includegraphics[scale=0.33]{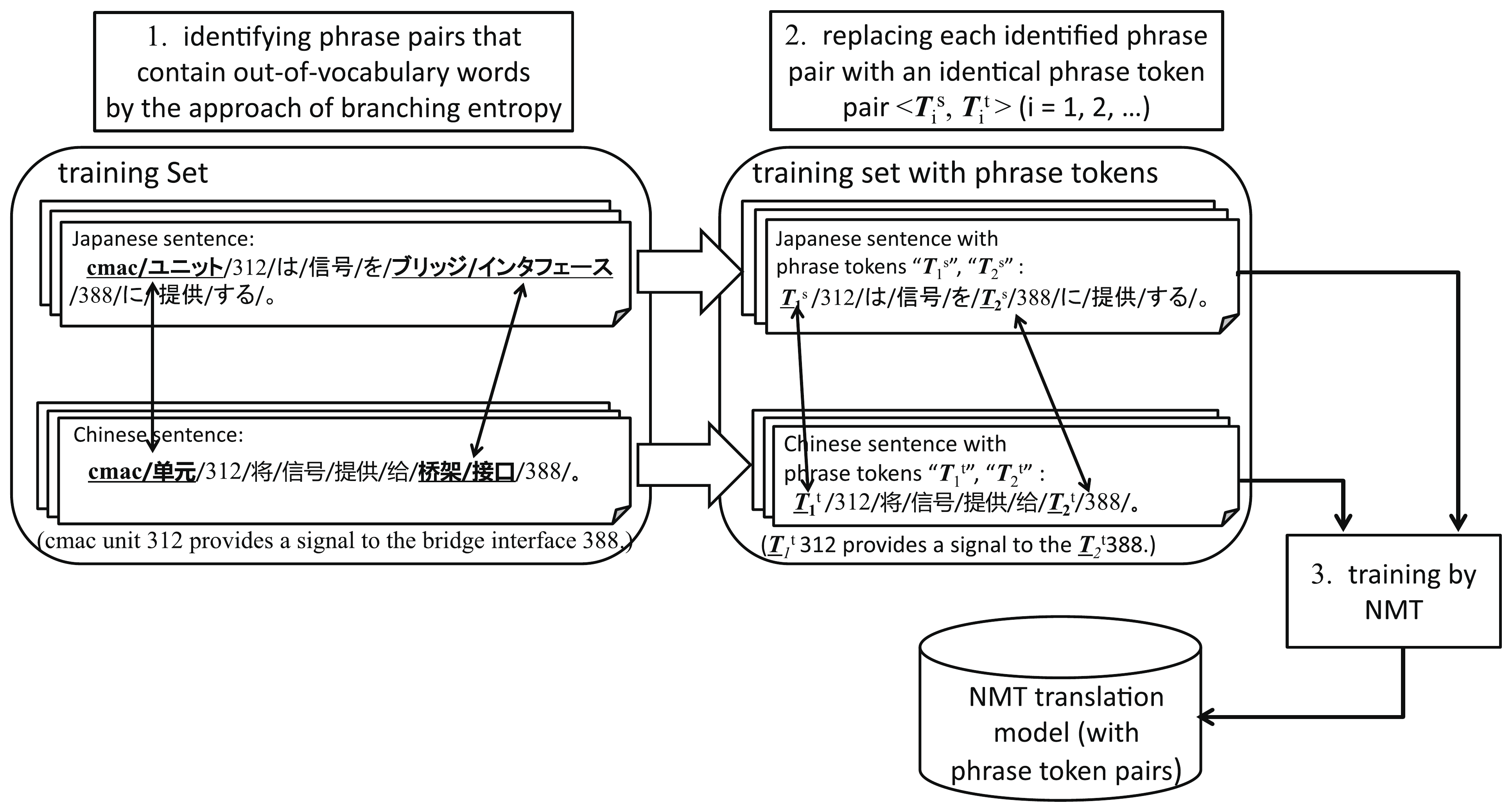}
   \caption{NMT training after replacing phrase pairs with
   token pairs $\langle T_{i}^s, T_{i}^t \rangle$ ($i=1,2,\ldots$)}
   \label{fig:train}
  \end{figure*}

\subsection{NMT Training after Replacing Phrase Pairs with Tokens}
\label{subsec:train}

  Figure~\ref{fig:train} illustrates the procedure for
  training the model with parallel
  patent sentence pairs in which phrase pairs are replaced with
  phrase token pairs $\langle T_1^s,T_1^t \rangle$, $\langle
  T_2^s,T_2^t\rangle$, and so on.

  In the step 1 of Figure~\ref{fig:train}, 
  source-target phrase pairs that contain at
  least one out-of-vocabulary word are selected from the training set using the
  branching entropy approach described in Section~\ref{sec:select_entr}.
  As shown in the step 2 of Figure~\ref{fig:train}, 
  in each of the parallel patent sentence pairs, occurrences
  of phrase pairs $\langle t_1^s,t_1^t \rangle$, $\langle
  t_2^s,t_2^t \rangle$, $\ldots$, $\langle t_k^s,t_k^t \rangle$ are then
  replaced with token pairs $\langle T_{1}^s,T_{1}^t \rangle$,
  $\langle T_{2}^s, T_{2}^t \rangle$, $\ldots$,
  $\langle T_{k}^s,T_{k}^t \rangle$. Phrase pairs $\langle t_1^{s},t_1^t \rangle$,
  $\langle t_2^s,t_2^t \rangle$, $\ldots$, $\langle t_k^s, t_k^t \rangle$ 
  are numbered in the order of occurrence of the source phrases 
  $t_1^{s}$ ($i=1,2,\ldots,k$) in each source sentence $S_s$. 
  Here note that in all the parallel sentence pairs $\langle S_s, S_t \rangle$, 
  the tokens pairs $\langle T_{1}^s,T_{1}^t \rangle$, $\langle
  T_{2}^s, T_{2}^t \rangle$, $\ldots$ that are
  identical throughout all the parallel sentence pairs are used in this procedure.
  Therefore, for example, in all the source patent sentences $S_s$, the phrase
  $t_1^{s}$ which appears earlier than other phrases 
  in $S_s$ is replaced with $T_{1}^s$.
  We then train the NMT model on a bilingual
  corpus, in which the phrase pairs are
  replaced by token pairs $\langle T_{i}^s, T_{i}^t \rangle$ ($i=1,2,\ldots$), 
  and obtain an NMT model in which the phrases
  are represented as tokens.\footnote{
   We treat the NMT system as a black box, and the strategy we present
   in this paper could be applied to any NMT
   system~\citep{Bahdanau15a,Cho14a,Kalch13a,Luong15b,Sutskever14a}. 
  }

\subsection{NMT Decoding and SMT Phrase Translation} 
\label{subsec:decode}

  \begin{figure*}
   \centering
   \includegraphics[scale=0.31]{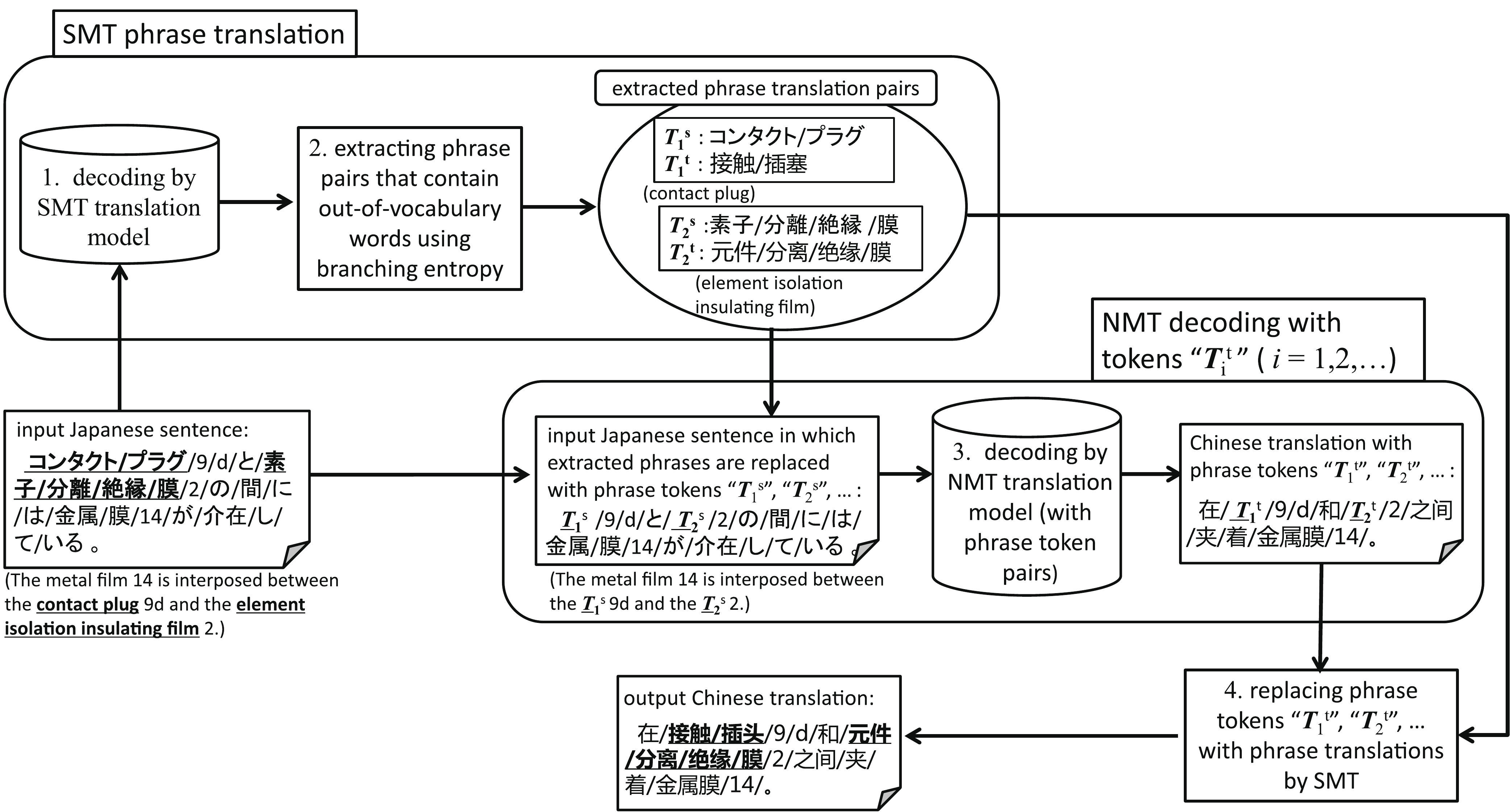}
     \caption{NMT decoding with tokens ``$T_{i}^s$''
   ($i=1,2,\ldots$) and the SMT phrase translation}
   \label{fig:decode}
  \end{figure*}

  Figure~\ref{fig:decode} illustrates the procedure for 
  producing target translations by decoding the input source sentence 
  using the method proposed in this paper.

  In the step 1 of Figure~\ref{fig:decode}, 
  when given an input source sentence, 
  we first generate its translation by decoding of SMT
  translation model.
  Next, as shown in the step 2 of Figure~\ref{fig:decode}, 
  we automatically extract the phrase pairs by branching
  entropy according to the procedure of
  Section~\ref{sec:select_entr}, where the input sentence
  and its SMT translation are considered as a pair of parallel sentence.
  Phrase pairs that contains at least one out-of-vocabulary word
  are extracted and are replaced with phrase token pairs $\langle
  T_{i}^s, T_{i}^t \rangle$ ($i=1,2,\ldots$). 
  Consequently, we have an input sentence in which the
  tokens ``$T_{i}^s$'' ($i=1,2,\ldots$) represent the
  positions of the phrases and a list of SMT phrase translations of
  extracted Japanese phrases. 
  Next, as shown in the step 3 of Figure~\ref{fig:decode}, 
  the source Japanese sentence with tokens is translated
  using the NMT model trained according to the procedure described in
  Section~\ref{subsec:train}.
  Finally, in the step 4, 
  we replace the tokens ``$T_{i}^t$'' ($i=1,2,\ldots$) of
  the target sentence translation with the phrase translations of the SMT.

\section{Evaluation}
\label{sec:exp}

\subsection{Patent Documents}
\label{sec:patent}
 
 Japanese-Chinese parallel patent documents were collected from the
 Japanese patent documents published
 by the Japanese Patent Office (JPO) during 2004-2012 and the Chinese
 patent documents published by the State
 Intellectual Property Office of the People's Republic of China (SIPO)
 during 2005-2010. From the collected documents, we
 extracted 312,492 patent families, and the method of \cite{Uchiyama07bs}
 was applied\footnote{
   Herein, we used a Japanese-Chinese translation lexicon comprising
   around 170,000 Chinese entries.
 } 
 to the text of the extracted patent families to align the Japanese and Chinese
 sentences. 
 The Japanese sentences were segmented into a sequence of morphemes using the
 Japanese morphological analyzer MeCab\footnote{
   \url{http://mecab.sourceforge.net/}
 }
 with the morpheme lexicon IPAdic,\footnote{
   \url{http://sourceforge.jp/projects/ipadic/}
 }
 and the Chinese sentences
 were segmented into a sequence of words using the Chinese
 morphological analyzer Stanford Word
 Segment~\citep{Tseng05a} trained using the Chinese Penn
 Treebank. 
 In this study, Japanese-Chinese parallel patent sentence
 pairs were ordered in descending order of sentence-alignment score
 and we used the topmost 2.8M pairs, 
 whose Japanese sentences contain fewer than 40 morphemes and
 Chinese sentences contain fewer than 40 words.\footnote{
   It is expected that the proposed NMT model can improve the baseline NMT
   without the proposed technique when translating longer sentences that
   contain more than 40 morphemes / words. It is because the approach
   of replacing phrases with tokens also shortens the input sentences,
   expected to contribute to solving the weakness of NMT model when
   translating long sentences.
 }

  Japanese-English patent documents are provided in the
  NTCIR-7 workshop~\citep{Fujii08c}, which are collected from the 10
  years of unexamined Japanese 
  patent applications published by the Japanese Patent Office (JPO) and
  the 10 years patent grant data published by the U.S. Patent \&
  Trademark Office (USPTO) in 1993-2000. The numbers of documents are
  approximately 3,500,000 for Japanese and 1,300,000 for
  English. From these document sets, patent families are 
  automatically extracted and the fields of ``Background of the
  Invention'' and ``Detailed Description of the Preferred Embodiments''
  are selected. Then, the method of \cite{Uchiyama07bs} is
  applied to the text of those fields, and Japanese and English
  sentences are aligned. The Japanese sentences were segmented into a sequence of
  morphemes using the Japanese morphological analyzer MeCab
  with the morpheme lexicon IPAdic.
  Similar to the case of Japanese-Chinese patent
  documents, in this study, out of the provided 1.8M Japanese-English
  parallel sentences, 1.1M parallel sentences 
  whose Japanese sentences contain fewer than 40 morphemes and
  English sentences contain fewer than 40 words are used.

  \begin{table*}
   \begin{center}
    \caption{Statistics of datasets}
    \label{tab:data_number}
    \begin{tabular}{|c||c|c|c|}
     \hline
       & training set & validation set & test set \\ \hline
     Japanese-Chinese & 2,877,178 & 1,000 & 1,000 \\ \hline
     Japanese-English & 1,167,198 & 1,000 & 1,000 \\ \hline
    \end{tabular}
   \end{center}
  \end{table*}

\subsection{Training and Test Sets}
\label{subsec:data}

  \begin{table*}
   \begin{center}
    {
    \caption{Automatic evaluation results (BLEU)}
    \label{tab:eva_result}
    \begin{tabular}{|l||c|c||c|c|}
     \hline
     System & ja $\rightarrow$ ch & ch $\rightarrow$ ja & ja
     $\rightarrow$ en & en $\rightarrow$ ja
     \\ \hline \hline

     Baseline SMT~\citep{Koehn07as} & 52.5 & 57.1 & 32.3 & 32.1
     \\ \hline 
     
     Baseline NMT & 56.5 & 62.5 & 39.9 & 41.5
     \\ \hline 

     NMT with PosUnk model & \multirow{2}{*}{56.9} &
	     \multirow{2}{*}{62.9} & \multirow{2}{*}{40.1} &
		     \multirow{2}{*}{41.9} \\ 
     \citep{Luong15a} &  &  &  &
     \\ \hline 

     NMT with phrase translation by SMT (phrase
     & \multirow{2}{*}{{\bf 57.7}} & \multirow{2}{*}{{\bf 64.2}} &
	     \multirow{2}{*}{{\bf 40.3}} & \multirow{2}{*}{{\bf 42.9}} \\
     pairs selected with branching entropy) & & & & \\ \hline
    \end{tabular}
}
   \end{center}
  \end{table*}

  We evaluated the effectiveness of the proposed NMT model at translating
  parallel patent sentences described in Section~\ref{sec:patent}. 
  Among the selected parallel sentence pairs, we randomly extracted
  1,000 sentence pairs for the test set and 1,000 sentence pairs for the
  validation set; the remaining sentence pairs were used for the
  training set. Table~\ref{tab:data_number} shows statistics of the datasets.

  According to the procedure of Section~\ref{sec:select_entr}, from the Japanese-Chinese
  sentence pairs of the training set, we collected 426,551 occurrences of
  Japanese-Chinese phrase pairs, which are 254,794 types of phrase pairs
  with 171,757 unique types of Japanese phrases and 129,071 unique
  types of Chinese phrases. Within the total 1,000 Japanese patent sentences in the
  Japanese-Chinese test set, 121 occurrences of Japanese phrases were
  extracted, which correspond to 
  120 types. With the total 1,000 Chinese patent sentences in the
  Japanese-Chinese test
  set, 130 occurrences of Chinese phrases were extracted, which
  correspond to 130 types.

  From the Japanese-English 
  sentence pairs of the training set, we collected 70,943 occurrences
  of Japanese-English phrase pairs, which are  61,017 types of phrase
  pairs with unique 57,675 types of Japanese phrases and 58,549 unique 
  types of English phrases.
  Within the total 1,000 Japanese patent sentences in the
  Japanese-English test set,
  59 occurrences of Japanese phrases were extracted, which correspond
  to 59 types. With the total 1,000 English patent sentences in the
  Japanese-English test
  set, 61 occurrences of English phrases were extracted, which
  correspond to 61 types.

\subsection{Training Details}
\label{subsec:detail}

  For the training of the SMT model, 
  including the word alignment and the phrase
  translation table, we used Moses~\citep{Koehn07as}, a toolkit for 
  phrase-based SMT models. We trained the SMT model on the training set
  and tuned it with the validation set.

  For the training of the NMT model, 
  our training procedure and hyperparameter choices were similar to those
  of \cite{Bahdanau15a}.
  The encoder consists of forward and
  backward deep LSTM neural networks each consisting of three
  layers, with 512 cells in each layer. The decoder is a three-layer
  deep LSTM with 512 cells in each layer. 
  Both the source vocabulary and the target
  vocabulary are limited to the 40K most-frequently used morphemes /
  words in the training set. 
  The size of the word embedding was set to 512.
  We ensured that all sentences in a minibatch were roughly the same
  length. Further training details are given below: 
  (1) We set the size of a minibatch to 128. 
  (2) All of the LSTM's parameter were initialized
  with a uniform distribution ranging between -0.06 and 0.06. 
  (3) We used the stochastic gradient descent, beginning at a fixed learning
  rate of 1. We trained our model for a total of 10 epochs, and we began
  to halve the learning rate every epoch after the first seven epochs. 
  (4) Similar to \cite{Sutskever14a}, we rescaled the
  normalized gradient to ensure that its norm does not exceed 5. 
  We trained the NMT model on the training set.
  The training time was around two days when using the described
  parameters on a 1-GPU machine.

  We compute the branching entropy using the frequency statistics from
  the training set.

  \begin{table*}
   \begin{center}
    \caption{Human evaluation results of pairwise evaluation
    (the score ranges from $-$100 to 100)}
    \label{tab:he_result_op}
    \begin{tabular}{|l||c|c||c|c|}
     \hline
     System & ja $\rightarrow$ ch & ch $\rightarrow$ ja & ja $\rightarrow$ en &
     en $\rightarrow$ ja
     \\ \hline \hline
      

      Baseline NMT & - & - & - & -
      \\ \hline

      NMT with PosUnk model & \multirow{2}{*}{9} &
	     \multirow{2}{*}{10.5} & \multirow{2}{*}{8} &
             \multirow{2}{*}{6.5} \\
      \citep{Luong15a} &  &  &  & 
      \\ \hline

       NMT with phrase translation by SMT (phrase
       & \multirow{2}{*}{{\bf 14.5}} & \multirow{2}{*}{{\bf 17}} 
 	    & \multirow{2}{*}{{\bf 11.5}} & \multirow{2}{*}{{\bf 15.5}}\\
     pairs selected with branching entropy) & & & &
       \\ \hline
    \end{tabular}
   \end{center}
  \end{table*}

  \begin{table*}
   \begin{center}
    \caption{Human evaluation results of JPO adequacy
    evaluation (the score ranges from 1 to 5)}
    \label{tab:he_result_jae}
    \begin{tabular}{|l||c|c||c|c|}
     \hline
     System & ja $\rightarrow$ ch & ch $\rightarrow$ ja & ja $\rightarrow$ en &
     en $\rightarrow$ ja
     \\ \hline \hline

     Baseline SMT~\citep{Koehn07as} & 3.5 & 3.7 & 3.1 & 3.2
		     \\ \hline

     Baseline NMT & 4.2 & 4.3 & 3.9 & 4.1
      \\ \hline

      NMT with PosUnk model & \multirow{2}{*}{4.3} &
	     \multirow{2}{*}{4.3} & \multirow{2}{*}{4.0} &
             \multirow{2}{*}{4.2} \\
      \citep{Luong15a} &  &  &  & 
      \\ \hline

       NMT with phrase translation by SMT (phrase
       & \multirow{2}{*}{{\bf 4.5}} & \multirow{2}{*}{{\bf 4.6}} 
 	    & \multirow{2}{*}{{\bf 4.1}} & \multirow{2}{*}{{\bf 4.4}}\\
     pairs selected with branching entropy)& & & & 
       \\ \hline
    \end{tabular}
   \end{center}
  \end{table*}
   
\subsection{Evaluation Results}
\label{subsec:eva}

  \begin{table*}
   \begin{center}
    {
    \caption{Numbers of untranslated morphemes / words of input sentences (for the test set)}
    \label{tab:err_num_unt}
    \begin{tabular}{|l||c|c||c|c|}
     \hline

     System & ja $\rightarrow$ ch & ch $\rightarrow$ ja & ja $\rightarrow$ en &
     en $\rightarrow$ ja
     \\ \hline \hline

	Baseline NMT & 89  & 92 & 415 & 226	     
      \\ \hline

       NMT with phrase translation by SMT (phrase
        & \multirow{2}{*}{43} & \multirow{2}{*}{45} &
		 \multirow{2}{*}{246} &  \multirow{2}{*}{134} \\
       pairs selected with branching entropy) & & & &
       \\ \hline
    \end{tabular}
}
   \end{center}
  \end{table*}

  In this work,  we calculated
  automatic evaluation scores for the translation results using a
  popular metrics called BLEU~\citep{papineni-EtAl:2002:ACL}.
  As shown in Table~\ref{tab:eva_result}, we report the
  evaluation scores, using the translations by
  Moses~\citep{Koehn07as} as the baseline SMT
  and the scores using the translations produced by the baseline NMT
  system without our proposed approach as the baseline NMT. 
  As shown in Table~\ref{tab:eva_result}, 
  the BLEU score obtained by the proposed NMT model is clearly 
  higher than those of the baselines. 
  Here, as described in Section~\ref{sec:extract}, the lower
  bounds of branching entropy for phrase
  pair selection are tuned as 5 throughout the evaluation of language pair of
  Japanese-Chinese, and tuned as 8 throughout the evaluation of language pair of
  Japanese-English, respectively.
  When compared with the baseline SMT, the 
  performance gains of the proposed system are approximately 5.2 BLEU
  points when translating Japanese into Chinese and 7.1 BLEU when
  translating Chinese into Japanese. 
  When compared with the baseline SMT, the
  performance gains of the proposed system are approximately 10.0
  BLEU points when translating Japanese into English and 10.8 BLEU when
  translating English into Japanese.
  When compared with the result of the baseline NMT,  
  the proposed NMT model achieved performance gains of 1.2 BLEU points
  on the task of translating Japanese into Chinese and 1.7 BLEU points on the
  task of translating Chinese into Japanese.
  When compared with the result of the baseline NMT,
  the proposed NMT model achieved performance gains of 0.4 BLEU points
  on the task of translating Japanese into English and 1.4 BLEU points
  on the task of translating English into Japanese.

  Furthermore, we quantitatively compared our study with
  the work of \cite{Luong15a}.
  Table~\ref{tab:eva_result} compares the NMT model with the PosUnk
  model, which is the best model proposed by \cite{Luong15a}. 
  The proposed NMT model achieves performance gains of 0.8 BLEU points
  when translating Japanese into Chinese, and performance gains
  of 1.3 BLEU points when translating Chinese into Japanese. 
  The proposed NMT model achieves performance gains of 0.2 BLEU points
  when translating Japanese into English, and performance gains
  of 1.0 BLEU points when translating English into Japanese.

  We also compared our study with the work of \cite{Long16b}.
  As reported in \cite{Long17arxiv}, when translating Japanese into
  Chinese, the BLEU of the NMT system of \cite{Long16b} in which all the selected compound nouns
  are replaced with tokens is 58.6, the BLEU of the NMT system in which
  only compound nouns that contain out-of-vocabulary 
  words are selected and replaced with tokens is 57.4, while the BLEU of
  the proposed NMT system of this paper is 57.7.
  Out of all the selected compound nouns of \cite{Long16b}, around 22\% contain
  out-of-vocabulary words, of which around 36\% share
  substrings with the phrases selected by branching entropy. The remaining
  78\% compound nouns do not contain out-of-vocabulary words and are considered to
  contribute to the improvement of BLEU points compared with the
  proposed method.  
  Based on this analysis, as one of our important future work, we
  revise the procedure in Section~\ref{sec:select_entr}
  of selecting phrases by branching entropy
  and then incorporate those in-vocabulary compound nouns
  into the set of the phrases selected by the branching entropy.

  \begin{figure*}
   \centering
   \includegraphics[scale=0.44]{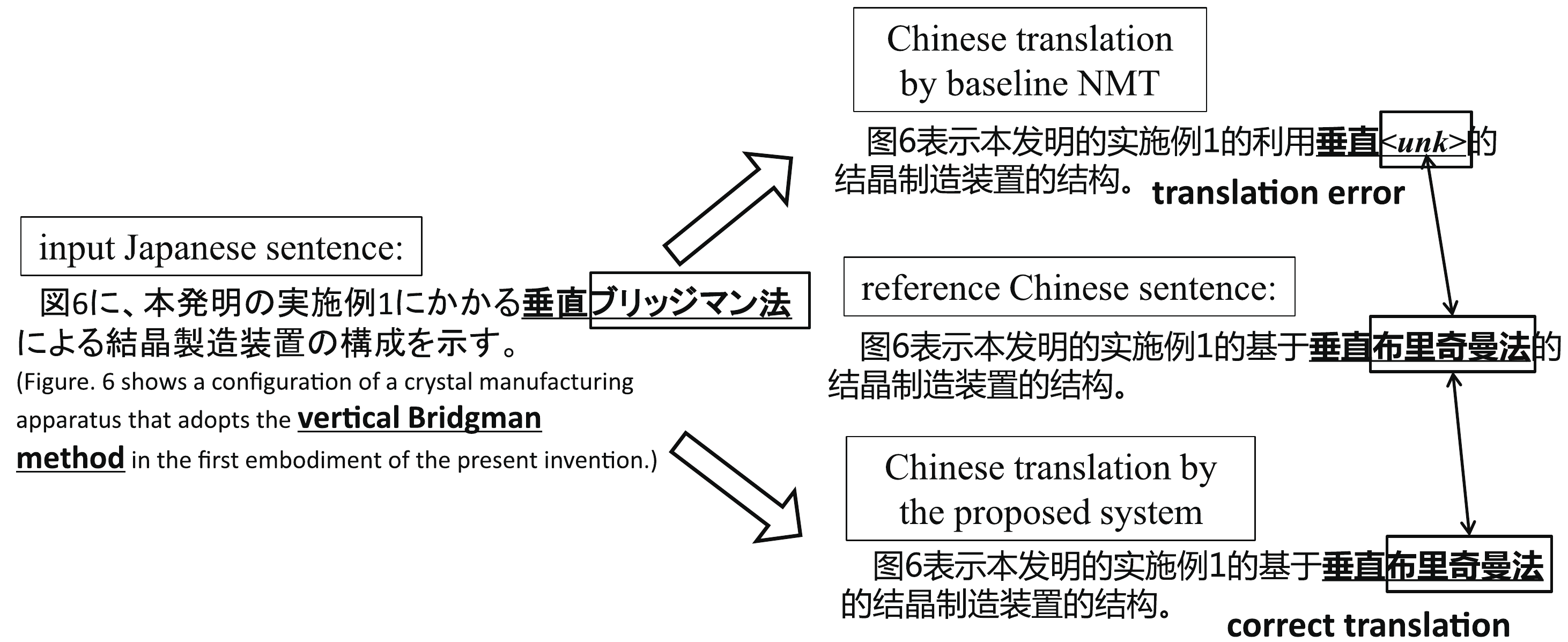}
     \caption{An example of correct translations produced by the proposed
   NMT model when addressing the problem of out-of-vocabulary words (Japanese-to-Chinese)}
   \label{fig:eva_correct}
   \vspace{-0.1cm}
  \end{figure*}
  
  \begin{figure*}[!tb]
   \centering
   \includegraphics[scale=0.44]{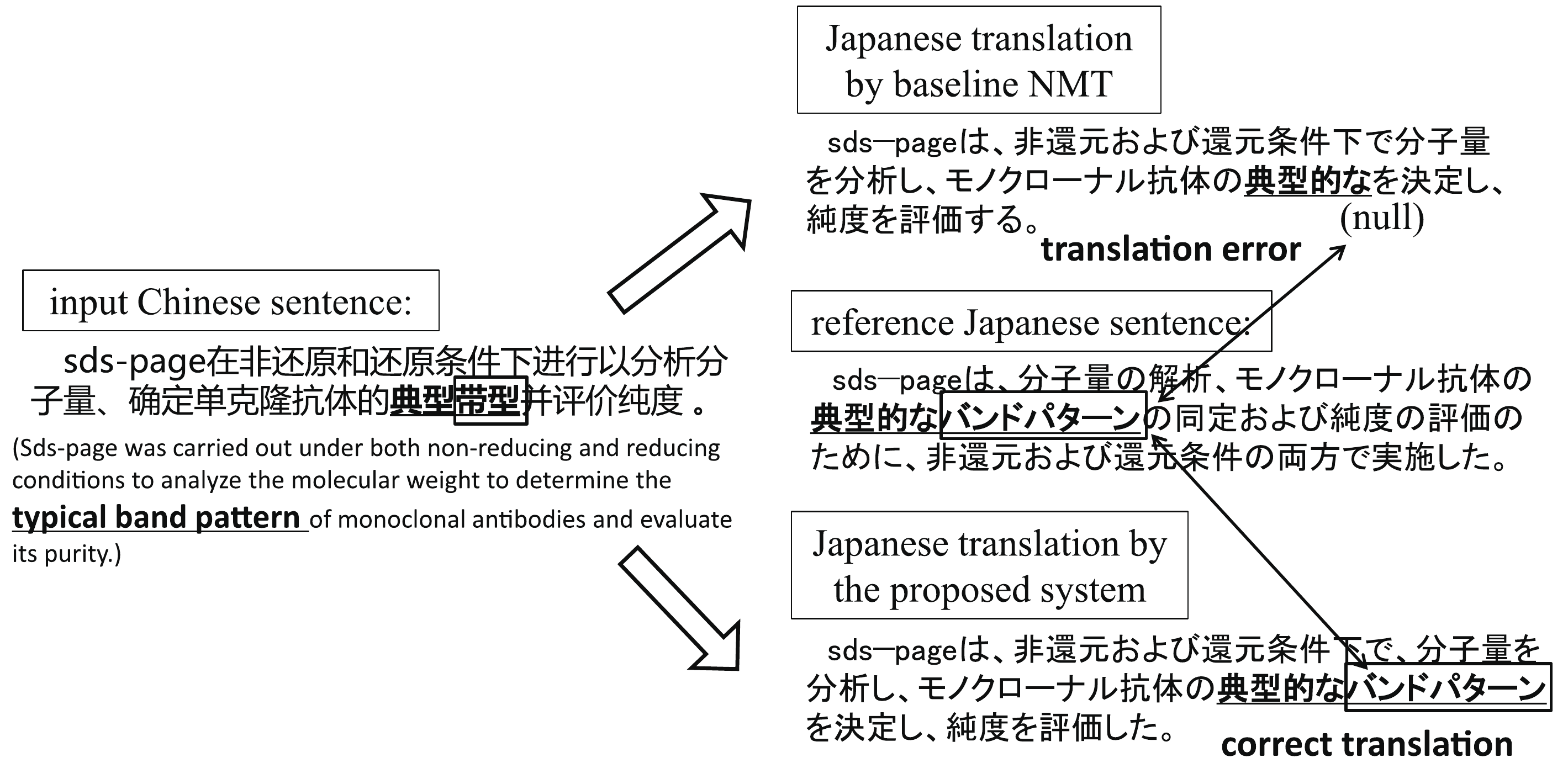}
     \caption{An example of correct translations produced by the proposed
   NMT model when addressing the problem of under-translation (Chinese-to-Japanese)}
   \label{fig:eva_correct_yaku}
   \vspace{-0.1cm}
  \end{figure*}

  In this study, we also conducted two types of human evaluations according to
  the work of \cite{Nakazawa15a}: 
  pairwise evaluation and JPO adequacy evaluation. 
  In the pairwise
  evaluation, we compared each translation produced by the baseline NMT
  with that produced by the proposed NMT model as well as the NMT model with PosUnk model,
  and judged which translation is better or whether they have comparable
  quality. The score of the pairwise evaluation is defined as below:
  \begin{eqnarray}
   score=100 \times \frac{W-L}{W+L+T} \nonumber
  \end{eqnarray}
  where W, L, and T are the numbers of translations that are better
  than, worse than, and comparable to the baseline NMT, respectively.
  The score of pairwise evaluation ranges from $-$100 to 100. 
  In the JPO adequacy evaluation, Chinese translations are evaluated according to the
  quality evaluation criterion for translated patent documents 
  proposed by the Japanese Patent Office (JPO).\footnote{
   \url{https://www.jpo.go.jp/shiryou/toushin/chousa/pdf/tokkyohonyaku_hyouka/01.pdf}
   (in Japanese)
  }
  The JPO adequacy criterion judges whether or not the technical factors and their relationships
  included in Japanese patent sentences are
  correctly translated into Chinese.
  The Chinese translations are then scored according to the percentage
  of correctly translated information, where a score
  of 5 means all of those information are translated
  correctly, while a score of 1 means that most 
  of those information are not translated correctly. 
  The score of the JPO adequacy evaluation is defined as the average over
  all the test sentences.
  In contrast to the study conducted by \cite{Nakazawa15a}, we randomly 
  selected 200 sentence pairs from the test set for human evaluation,
  and both human evaluations were conducted using only one
  judgement. Table~\ref{tab:he_result_op} and
  Table~\ref{tab:he_result_jae} shows the results of the human evaluation for the
  baseline SMT, baseline NMT, NMT model with PosUnk model, and the proposed
  NMT model. We observe that the proposed model achieves the best
  performance for both the pairwise and JPO adequacy evaluations when we
  replace the tokens with SMT phrase translations after decoding the 
  source sentence with the tokens.

  \begin{figure*}
   \centering
   \includegraphics[scale=0.43]{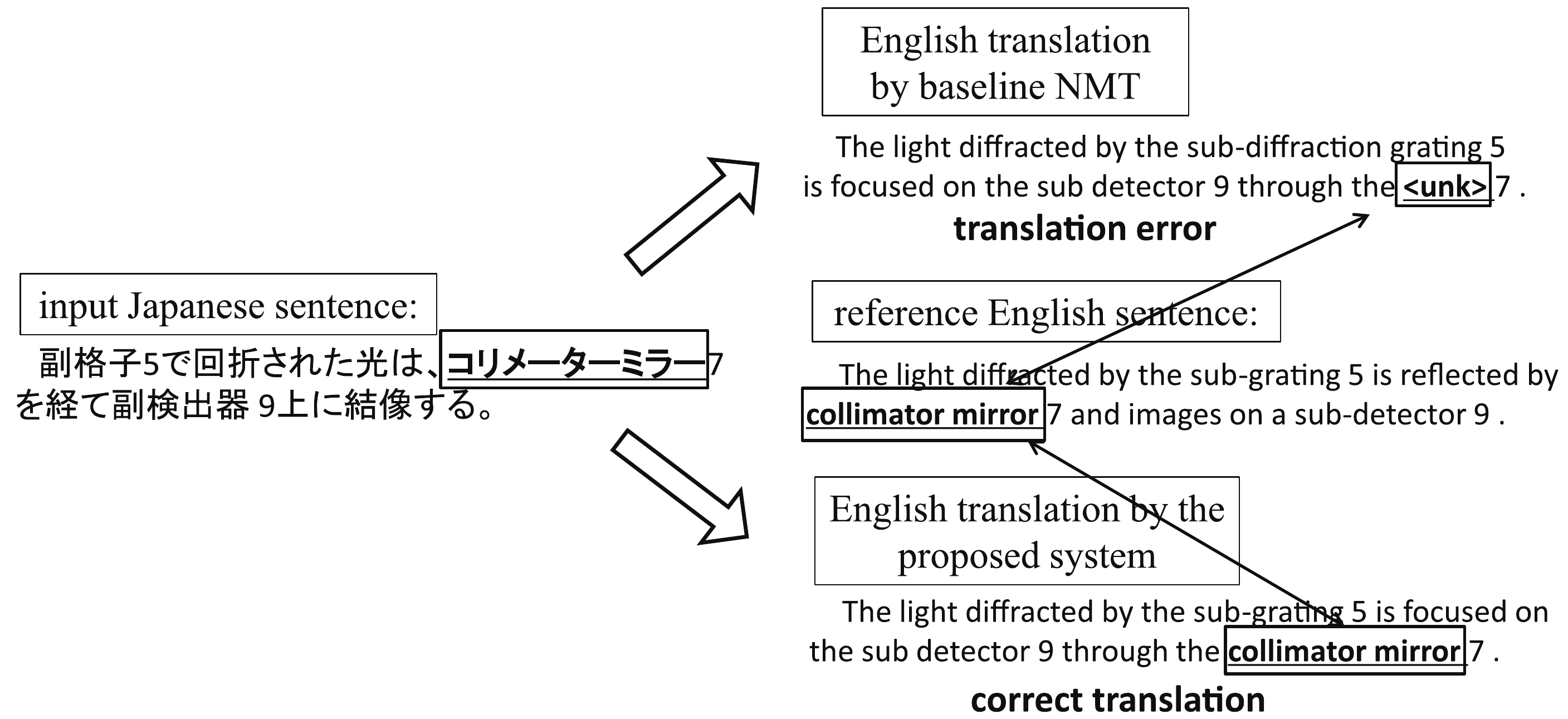}
     \caption{An example of correct translations produced by the proposed
   NMT model when addressing the problem of out-of-vocabulary words (Japanese-to-English)}
   \label{fig:eva_correct_je}
   \vspace{-0.2cm}
  \end{figure*}
  
  \begin{figure*}[!tb]
   \centering
   \includegraphics[scale=0.43]{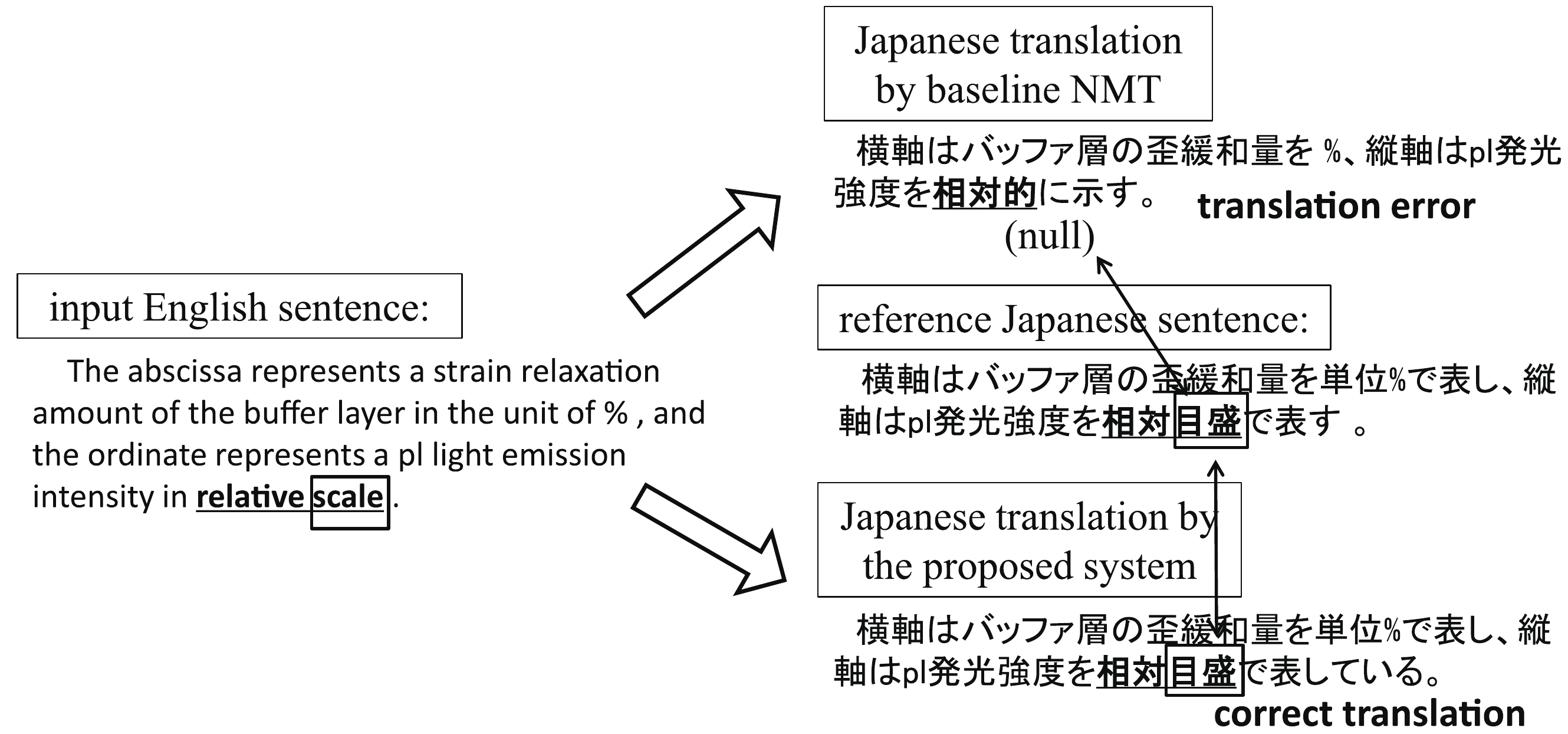}
     \caption{An example of correct translations produced by the proposed
   NMT model when addressing the problem of under-translation (English-to-Japanese)}
   \label{fig:eva_correct_yaku_je}
   \vspace{-0.2cm}
  \end{figure*}

  For the test set, we also counted the numbers of 
  the untranslated words of input sentences.
  As shown in Table~\ref{tab:err_num_unt}, 
  the number of untranslated words by the
  baseline NMT reduced to around 50\% in the cases of ja $\rightarrow$ ch and ch
  $\rightarrow$ ja by the proposed NMT model, and reduced to around
  60\% in the cases of ja $\rightarrow$ en and en $\rightarrow$ 
  ja.\footnote{
  Although we omit the detail of the evaluation results of untranslated
  words of the NMT model with PosUnk model~\citep{Luong15a} in
  Table~\ref{tab:err_num_unt}, the number 
  of the untranslated words of the NMT model with PosUnk 
  model is almost the same as that of the baseline NMT, which is much more
  than that of the proposed NMT model.
  }
  \footnote{
  Following the result of an additional evaluation where having
  approximately similar size of the training parallel sentences
  between the language pairs of Japanese-to-Chinese/Chinese-to-Japanese
  and Japanese-to-English/English-to-Japanese,
  we concluded that the primary reason why the numbers of untranslated
  morphemes / words tend to be much larger in the case of the language pair
  of Japanese-to-English/English-to-Japanese
  than in the case of the language pair of
  Japanese-to-Chinese/Chinese-to-Japanese
  is simply the matter of a language specific issue.
  }
  This is mainly because part of untranslated source words are out-of-vocabulary,
  and thus are untranslated by the baseline NMT. The proposed system
  extracts those out-of-vocabulary words as a part of phrases and replaces
  those phrases with tokens before the decoding of NMT. Those phrases
  are then translated by SMT and inserted in the output translation,
  which ensures that those out-of-vocabulary words are translated.     

  Figure~\ref{fig:eva_correct} compares an example of correct translation
  produced by the proposed system with one produced by the
  baseline NMT. In this example, the translation is a translation error
  because the Japanese word
  ``\includegraphics[scale=0.45]{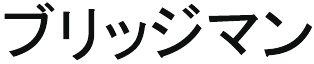} (Bridgman)''
  is an out-of-vocabulary word and is erroneously translated into the
  ``$\langle unk \rangle$'' token. The proposed NMT model correctly translated the Japanese
  sentence into Chinese, where the out-of-vocabulary word
  ``\includegraphics[scale=0.45]{figures/bridgman_ja.eps}'' is correctly
  selected by the approach of branching entropy as a part of the Japanese
  phrase ``\includegraphics[scale=0.45]{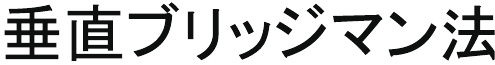}
  (vertical Bridgman method)''. The selected Japanese
  phrase is then translated by the phrase translation table
  of SMT. 
  Figure~\ref{fig:eva_correct_yaku} shows another example of correct
  translation produced by the proposed system with one produced by the
  baseline NMT. As shown in Figure~\ref{fig:eva_correct_yaku}, the
  translation produced by baseline NMT is a translation error
  because the out-of-vocabulary Chinese word
  ``\includegraphics[scale=0.45]{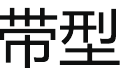} (band
  pattern)'' is an untranslated word and its translation is not
  contained in the output translation of the baseline NMT. The  
  proposed NMT model correctly translated the Chinese word into
  Japanese because the Chinese word
  ``\includegraphics[scale=0.45]{figures/bandpattern_ch.eps} (band
  pattern)''is selected as a part of Chinese phrase
  ``\includegraphics[scale=0.45]{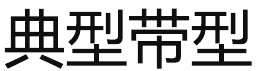}(typical
  band pattern)'' with branching entropy and then is translated by SMT.
  Moreover, Figure~\ref{fig:eva_correct_je} and
  Figure~\ref{fig:eva_correct_yaku_je} compare examples of 
  correct translations produced by the proposed system with those produced
  by the baseline NMT when translating patent sentences in both directions
  of Japanese-to-English and English-to-Japanese.

\section{Conclusion}
\label{sec:fin}

  This paper proposed selecting phrases that contain
  out-of-vocabulary words using the branching
  entropy. These selected phrases are then replaced with tokens
  and post-translated using an SMT phrase translation. Compared with the
  method of \cite{Long16b}, the contribution of the proposed NMT model
  is that it can be used on any language pair without language-specific
  knowledge for technical terms selection. 
  We observed that the proposed NMT model performed much better than the baseline NMT
  system in all of the language pairs:
  Japanese-to-Chinese/Chinese-to-Japanese and Japanese-to-English/English-to-Japanese.
  One of our important future tasks is to compare the translation
  performance of the proposed NMT model with that based on subword units
  (e.g. \cite{Sennrich16a}).
  Another future task is to improve the performance of the present
  study by incorporating the in-vocabulary non-compositional phrases,
  whose translations cannot be obtained by translating their constituent
  words. It is expected to achieve a better translation performance by
  translating those kinds of phrases using a phrase-based SMT instead of
  using NMT. 

\bibliographystyle{apalike}
\bibliography{myabbrv,mydb,long}

\end{document}